\pdfoutput=1
\documentclass[11pt]{article}

\usepackage{acl}

\usepackage{tabularx}
\usepackage{booktabs}
\usepackage{siunitx}
\usepackage{colortbl}
\usepackage{xcolor}
\usepackage{multicol}
\usepackage{multirow}

\usepackage{enumitem}

\usepackage{pgfplots}
\pgfplotsset{compat=1.18}

\usepackage[most]{tcolorbox}

\newtcolorbox{promptbox}{
    colback=black!5,      % Background color
    colframe=black!75,    % Frame color
    width=\textwidth,     % Box width
    arc=2mm,              % Corner radius
    boxrule=0.5pt,        % Frame thickness
}
\newtcolorbox{promptcaption}{
    colback=white,
    colframe=white,
    fontupper=\small,
    width=\textwidth,     % Box width
    arc=0mm,              % Corner radius
    boxrule=0pt,        % Frame thickness
    top=2pt,     % Sets the top padding
    bottom=2pt,   % Sets the bottom padding
    halign=center,
}

\usepackage{hyperref}

\usepackage{pifont}
\newcommand{\cmark}{\textcolor{green}{\ding{51}}}%
\newcommand{\xmark}{\textcolor{red}{\ding{55}}}%

\usepackage{times}
\usepackage{latexsym}

\usepackage[T1]{fontenc}
\usepackage[utf8]{inputenc}

\usepackage{microtype}

\usepackage{inconsolata}

\usepackage{graphicx}

\definecolor{mygreen}{HTML}{2E7D32}      % A nice, dark green for text
\definecolor{mygreenbg}{HTML}{E8F5E9}    % A very light green for background
\definecolor{myorange}{HTML}{E87400}    % A strong orange for text
\definecolor{myorangebg}{HTML}{FEF3DE}  % A light yellow/orange for background
\definecolor{mygray}{HTML}{546E7A}      % A neutral, dark gray for text
\definecolor{mygraybg}{HTML}{CFD8DC}    % A light gray for background

\newcommand{\diffbox}[4]{%
    \small % Use a slightly smaller font inside the box
    \tikz[baseline=(char.base)]\node[
        rounded corners=3pt,    % Rounded edges
        fill=#1,                % Background color
        text=#2,                % Text color
        inner sep=2pt,          % Padding inside the box
        font=\bfseries,         % Bold text
        ] (char) {#3 #4};%
}

\newcommand{\goodup}[1]{\diffbox{mygreenbg}{mygreen}{$\uparrow$}{#1}}
\newcommand{\gooddown}[1]{\diffbox{mygreenbg}{mygreen}{$\downarrow$}{#1}}
\newcommand{\badup}[1]{\diffbox{myorangebg}{myorange}{$\uparrow$}{#1}}
\newcommand{\baddown}[1]{\diffbox{myorangebg}{myorange}{$\downarrow$}{#1}}
\newcommand{\neutral}[1]{\diffbox{mygraybg}{mygray}{$\rightarrow$}{#1}}
\usepackage{cleveref}
\crefformat{section}{\S#2#1#3} % see manual of cleveref, section 8.2.1
\crefformat{subsection}{\S#2#1#3}
\crefformat{subsubsection}{\S#2#1#3}

\title{OpenWHO: A Document-Level Parallel Corpus for Health Translation in Low-Resource Languages}
\author{Raphaël Merx$^\lambda$ \hspace{0.5cm} Hanna Suominen$^{\psi,\, \phi}$ \hspace{0.5cm} {\bf Trevor Cohn$^{\lambda}$} \hspace{0.5cm} {\bf Ekaterina Vylomova$^\lambda$} \\
$^\lambda$ School of Computing and Information Systems, The University of Melbourne \\ $^\psi$School of Computing, The Australian National University \\
$^\phi$School of Medicine and Psychology, The Australian National University
}

\begin{document}
\maketitle

\begin{abstract}
In machine translation (MT), health is a high-stakes domain characterised by widespread deployment and domain-specific vocabulary. However, there is a lack of MT evaluation datasets for low-resource languages in this domain. To address this gap, we introduce OpenWHO, a document-level parallel corpus of 2,978 documents and 26,824 sentences from the World Health Organization's e-learning platform. Sourced from expert-authored, professionally translated materials shielded from web-crawling, OpenWHO spans a diverse range of over 20 languages, of which nine are low-resource. Leveraging this new resource, we evaluate modern large language models (LLMs) against traditional MT models. Our findings reveal that LLMs consistently outperform traditional MT models, with Gemini 2.5 Flash achieving a +4.79 ChrF point improvement over NLLB-54B on our low-resource test set. Further, we investigate how LLM context utilisation affects accuracy, finding that the benefits of document-level translation are most pronounced in specialised domains like health. We release the OpenWHO corpus to encourage further research into low-resource MT in the health domain.
\end{abstract}

\section{Introduction}

Translation in the health domain combines clinical risks, widespread demand, and domain-specific complexity \cite{mehandru_reliable_2022,neves-etal-2024-findings}. By offering a timely and resource-efficient complement to human translation, machine translation (MT) can lower the barrier to disseminating health content, from education materials for local health workers \cite{hammond_lessons_2024} to public safety information during crises \cite{federici_ethics_2023,utunen_scale_2023}. However, evaluation of MT in the health domain is hampered by a lack of datasets that cover a wide range of languages, particularly low-resource ones. Notable exceptions include the TICO-19 corpus, which focuses on COVID-19 \cite{anastasopoulos-etal-2020-tico}, and the \textsc{AfriDoc-MT} corpus, which covers health articles for five African languages \cite{alabi_afridoc-mt_2025}. Nonetheless, a need remains for evaluation datasets sourced from expert-authored, professionally translated materials that span a wide and diverse set of low-resource languages.

% source: https://docs.google.com/drawings/d/1ARcrSU1rw_lIf4Q63ixOuKKOMQ3W1QpgzFDKZJkJj3I/edit
% low-res scripts: Armenian, Georgian, Sinhala
\begin{figure}
    \centering
    \includegraphics[width=0.9\linewidth]{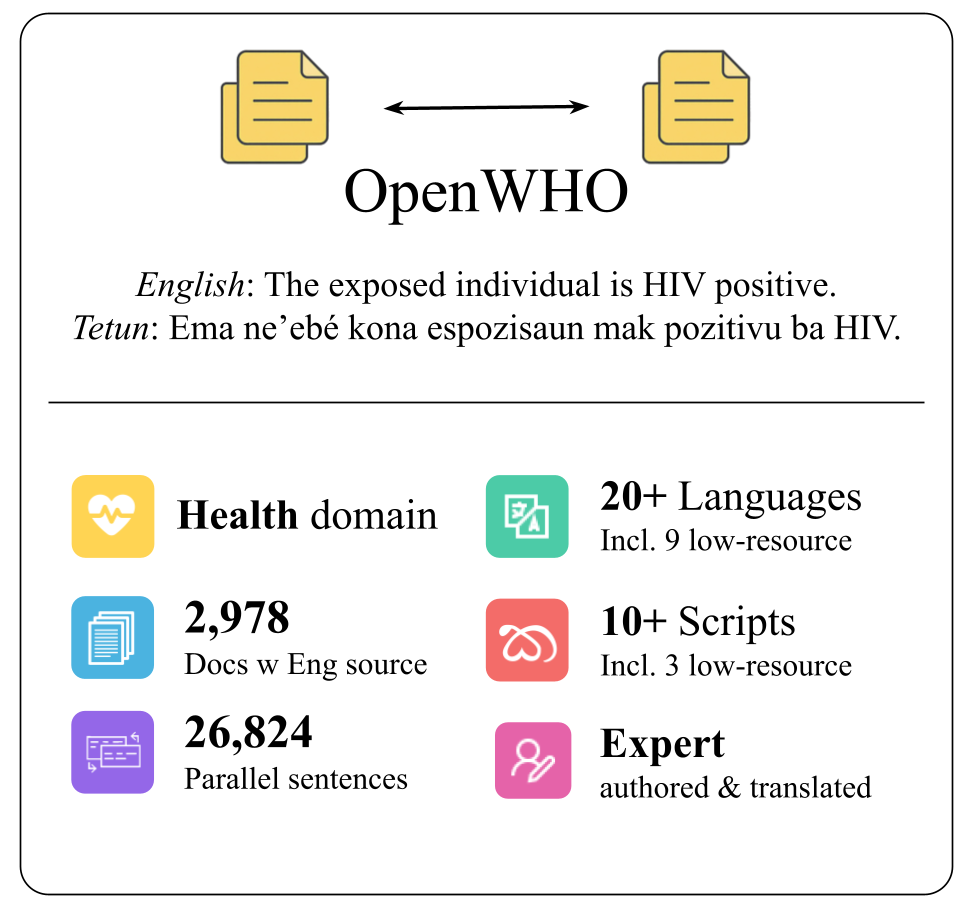}
    \caption{Overview of the OpenWHO parallel dataset, highlighting its depth across low-resource languages and scripts.}
    \label{fig:figure-1}
\end{figure}

\begin{figure*}[ht]
\centering
\begin{tikzpicture}
    % --- AXIS (Sentences) ---
    \begin{axis}[
        width=0.9\textwidth,
        height=5cm,
        ybar,                 % Style for bar chart
        bar width=6pt,        % Width of each bar
        ylabel={\textbf{\# sentences}},
        ylabel style={font=\small},
        enlarge x limits=0.03,
        legend style={at={(0.5,-0.3)}, anchor=north, legend columns=-1}, % Legend at bottom
        ymode=log,
        yticklabel style={/pgf/number format/fixed}, % Prevents scientific notation
        % Define the x-axis ticks and labels in descending order of sentences
        symbolic x coords={
            Macedonian, Arabic, French, Russian, Georgian, Armenian, Azerbaijani, Ukrainian, Turkish,
            Tetun, Dutch, Albanian, Kazakh, Chinese, Spanish, Indonesian, Portuguese, Somali,
            Sinhala, Tamil, Persian, Hindi, Amharic, Marathi
        },
        xtick=data,
        xticklabels={
            \textbf{Macedonian}, Arabic, French, Russian, \textbf{Georgian}, \textbf{Armenian}, \textbf{Azerbaijani}, Ukrainian, Turkish,
            \textbf{Tetun}, Dutch, \textbf{Albanian}, \textbf{Kazakh}, Chinese, Spanish, Indonesian, Portuguese, \textbf{Somali},
            \textbf{Sinhala}, Tamil, Persian, Hindi, Amharic, Marathi
        },
        xticklabel style={
            rotate=45,
            anchor=east,
            font=\small
        },
    ]
    % Data for the bar plot (Sentences), sorted from most to least
    \addplot[fill=red!70, draw=red!90!black] coordinates {
        (Macedonian,3695)
        (Arabic,2623)
        (French,2385)
        (Russian,2194)
        (Georgian,2151)
        (Armenian,1982)
        (Azerbaijani,1677)
        (Ukrainian,1632)
        (Turkish,1093)
        (Tetun,1086)
        (Dutch,1082)
        (Albanian,1029)
        (Kazakh,936)
        (Chinese,871)
        (Spanish,596)
        (Indonesian,329)
        (Portuguese,279)
        (Somali,224)
        (Sinhala,214)
        (Tamil,207)
        (Persian,56)
        (Hindi,35)
        (Amharic,25)
        (Marathi,19)
    };
    \end{axis}
\end{tikzpicture}
\caption{Number of parallel sentences per language in the OpenWHO dataset. The English source has 50,898 sentences. Low-resource languages covered in our experiments (Section~\ref{sec:experiments}) are in \textbf{bold}.}
\label{fig:lang_stats_sentences_sorted}
\end{figure*}
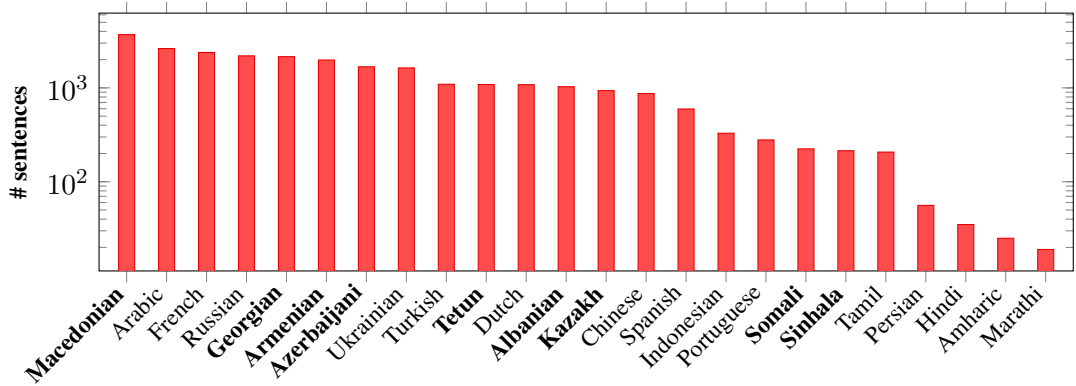

To address this gap, we introduce OpenWHO, a document-level parallel corpus designed for evaluating health MT. Sourced from the World Health Organization's multilingual e-learning platform, its content is expert-authored, professionally translated, and shielded from web-crawling, thus minimising contamination risk. The corpus covers over 20 languages, nine of which are low-resource, including some with low-resource scripts like Armenian, Georgian, and Sinhala. By focusing on health education, a domain fundamental to local quality of care \cite{merx_machine_2024}, OpenWHO provides a realistic benchmark for a high-impact MT use case.

% INTRO: option 1

Leveraging this new resource, we conduct a systematic evaluation comparing modern large language models (LLMs) against traditional NMT systems. For LLMs, we study different context strategies (document-level, sentence-level, etc) and to determine whether the benefits of document-level translation are specific to our dataset, we extend our evaluation to the news and literary subsets of the WMT24++ benchmark \cite{deutsch_wmt24_2025}.

Our main contributions are \footnote{Code: \href{https://github.com/raphaelmerx/openwho-code}{github.com/raphaelmerx/openwho-code} \\ Dataset: \href{https://huggingface.co/datasets/raphaelmerx/openwho}{huggingface.co/datasets/raphaelmerx/openwho}}:
\begin{itemize}
    \item \textbf{We introduce and release OpenWHO}, a parallel corpus for health MT that covers low-resource languages. It comprises 2,978 documents and 26,824 parallel sentences from expert-authored, professionally translated materials  (\cref{sec:openwho-dataset}).
    \item \textbf{We show that modern LLMs outperform traditional NMT models} for low-resource translation in the health domain. Our findings show Gemini 2.5 Flash with document-level context achieves a +4.79 ChrF point improvement over NLLB-54B on our test set (\cref{sec:exp-results}).
    \item \textbf{We find that the benefit of document-level context is model and domain-dependent} for low-resource MT. Accuracy gains are most pronounced when using best-performing models to translate specialised domains like health and literature, while the general (news) domain shows more modest improvements, highlighting that domain complexity drives context utility (\cref{sec:exp-results}).
\end{itemize}

\begin{table*}[t]
\small
\centering
\begin{tabular}{l c c c c}
\toprule
\textbf{Work} & \textbf{Low-resource} & \textbf{LLMs} & \textbf{Document-level} & \textbf{Specialised domains} \\
\midrule
Ours & \cmark & \cmark & \cmark & \cmark \\
\midrule
\citet{post_escaping_2024} & \xmark & \xmark & \cmark & \xmark \\
\citet{wang_document-level_2023}       & \xmark & \xmark & \cmark & \xmark \\
\citet{koneru_contextual_2024}     & \xmark & \cmark & \cmark & \xmark \\
\citet{karpinska_large_2023}  & \xmark & \cmark & \cmark & \cmark \\
\citet{enis_llm_2024}       & \cmark & \cmark & \cmark & \xmark \\
\citet{zebaze_compositional_2025}     & \cmark & \cmark & \xmark & \cmark \\
\citet{yang_optimising_2024} & \xmark & \cmark & \xmark & \cmark \\
\citet{mohammed_context-aware_2025} & \xmark & \cmark & \cmark & \xmark \\
\citet{pang_salute_2025} & \xmark & \cmark & \cmark & \cmark \\
\bottomrule
\end{tabular}
\caption{Comparison with prior work on context utilisation for MT.}
\label{tab:related_work_summary}
\end{table*}

\section{Related Work}

\paragraph{Document-level MT}
Document-level MT has long been recognised as desirable, as it allows models to leverage broader discourse for improved coherence and accuracy \cite{maruf_survey_2021}. Early work with traditional NMT models has shown mixed results, with some studies demonstrating that document-level context can significantly improve translation quality \cite{miculicich-etal-2018-document,wang_document-level_2023,post_escaping_2024}, while others questioned whether these improvements stemmed from true contextual understanding, arguing that the context encoder was not modeling discourse but acting as a "noise generator" that improves model robustness, rather than leveraging discourse information \cite{li-etal-2020-multi-encoder,appicharla-etal-2024-case}.

\paragraph{LLMs for document-level MT}
LLMs, given their ability to process extended contexts, are well placed to benefit from document-level context. \citet{koneru_contextual_2024} explored contextual translation with Llama2-13B on English-German, finding mixed results, where document-level context sometimes providing no performance gains over sentence-level translation. \citet{karpinska_large_2023} demonstrated that paragraph-level translation outperforms sentence-level approaches in literary fiction using GPT-3.5, though they have noted that their findings might not generalise to low-resource settings. \citet{yang_optimising_2024} fine-tuned LLaMA3-8B for context-aware translation, showing that increasing context window size yields gains, particularly when evaluated with neural metrics. Recent mechanistic analysis by \citet{mohammed_context-aware_2025} revealed that LLMs can be surprisingly ``context-insensitive,'' with smaller models showing limited ability to effectively utilise available context, a finding that may explain varying performance observed in earlier works.

\paragraph{Low-resource MT with LLMs} While LLMs have shown promising results on low-resource MT \cite{guo-etal-2024-teaching,merx-etal-2024-low}, evaluation has been predominantly limited to sentence-level translation. \citet{enis_llm_2024} demonstrated that Claude significantly outperforms NLLB on Yoruba-English translation, finding substantial improvements from document-level over sentence-level translation, though their evaluation focused solely on the low-resource-to-English direction. \citet{zebaze_compositional_2025} explored low-resource translation with LLMs using compositional approaches on datasets including FLORES, NTREX, and TICO-19, but operated at the sentence level with few-shot learning rather than document-level context. This sentence-level focus in low-resource settings represents a significant gap, as the dynamics of context utilisation may be fundamentally different for low-resource languages where models have seen limited training data and where translation errors could compound across sentences within a document.

Gaps remain in our understanding of document-level translation for low-resource languages in specialised domains. First, there is a shortage of evaluation data for document-level low-resource machine translation in specialised domains, such as healthcare. Second, there has been no systematic analysis of how LLMs utilise document-level context when translating low-resource languages, particularly in specialised domains where coherence and terminological consistency are required.

\section{The OpenWHO dataset}
\label{sec:openwho-dataset}

\subsection{Source and Motivation}

% - why it's high-q: content authored and vetted by WHO experts, authoritative and accurate information, and a deliberate multilingual strategy

\paragraph{The OpenWHO platform.} Our corpus is drawn from OpenWHO.org, the World Health Organization's (WHO) former e-learning platform for public health education. Active from 2017 to 2024, the platform's primary goal was to disseminate health knowledge to healthcare professionals, frontline responders, and the public, particularly during health emergencies \cite{george_ensuring_2022, utunen_observations_2023}. The content was authored and vetted by WHO experts and its global network of partner institutions, ensuring that the information and its translations were authoritative, accurate, and reflected up-to-date scientific guidance \cite{george_ensuring_2022}. The topics covered a wide range of public health issues, including specific disease responses (e.g., COVID-19, Ebola), vaccination protocols, infection prevention, and emergency preparedness \cite{utunen_global_2020, utunen_observations_2023}.

\paragraph{Multilingual focus.} A key tenet of the OpenWHO initiative was to ensure equitable access to information, which included a deliberate strategy of multilingual dissemination. Course materials were translated from English into a range of languages, with a focus on providing resources for low- and middle-income countries \cite{george_ensuring_2022, utunen_observations_2023}. The course-based format ceased operations in December 2024, transitioning to a static resource library. The data for our corpus was collected prior to this change and exclusively comprises materials from the course-based period (2017–2024). This commitment to creating expert-authored, multilingual content made the OpenWHO platform a high-quality source for extracting a document-level parallel corpus in the health domain, covering several low-resource~languages.

Because all course material was hosted behind a login screen, it was shielded from the large-scale web crawling that constitutes the training data for most LLMs, mitigating risk of pre-training contamination. To confirm this, we conducted searches across publicly available web-scraped corpora (C4, MADLAD), and performed targeted web searches (via Google Search) using sentences found in OpenWHO course content. These searches revealed only publicly accessible OpenWHO course descriptions (which are not part of our corpus), with no course content found within these data sources.

\begin{table*}[ht]
\small
\centering
\begin{tabularx}{\linewidth}{>{\raggedright\hsize=0.6\hsize}X >{\hsize=1.4\hsize}X}
\toprule
\textbf{Strategy} & \textbf{Description} \\
\midrule
\textbf{Sentence level} & Our baseline. Each sentence is translated independently without any additional context.\\
\addlinespace
\textbf{Sentence window} (batched sliding window in \citet{koneru_contextual_2024}) & A constrained-context approach. The model receives only the immediately preceding and succeeding sentences as context, aiming to capture local discourse phenomena without overwhelming the model. \\
\addlinespace
% Sent + doc context (+) model gets lots context (+) model has less opportunity to trip over itself (hallucinations etc) (-) model doesn't get any context in the TGT language
\textbf{Sentence + doc context} & The model is provided with the full source document as context within the prompt but is instructed to translate only the single, target sentence. \\
\addlinespace
% Doc level (+) model gets lots context (+) model can generate TGT text and keep it consistent with itself (-) model can trip over itself and hallucinate
\textbf{Document level} & The model is given the entire source document and instructed to translate the whole text. As per \citet{enis_llm_2024}, we use one sentence per line, and evaluate at the sentence level after translation.  \\
\addlinespace
\textbf{Doc-level + self-correct}, as per \cite{wu_please_2025} & A two-step approach: (1) Document-level translation then (2) feed the generated translation back to the model with a new prompt asking it to review and improve its own output, testing its self-correction and refinement capabilities. \\
\bottomrule
\end{tabularx}
\caption{The five translation strategies evaluated in our experiments. Each strategy represents a different approach to leveraging context for machine translation. Associated prompts are in Appendix~\ref{sec:appendix-prompts}.}
\label{tab:translation_strategies}
\end{table*}

\subsection{Data Curation Pipeline}

\subsubsection{Document Extraction}

\paragraph{Scraping}
While we secured authorization from the WHO to collect and release this data, a direct database export was not available. Therefore, in consultation with the WHO, we developed a web scraping pipeline to gather the course materials. Using the Scrapy framework,\footnote{\href{https://www.scrapy.org/}{https://www.scrapy.org/}} we developed a web scraper to navigate the OpenWHO site, enrol in each individual course, and extract the raw HTML content of every course page. Each page was uniquely identified by its course ID and language, as well as its position within the course structure (section and subsection numbers).

\paragraph{Content filtering.} A significant portion of the OpenWHO curriculum relies on video-based learning. As our focus is on creating a parallel text corpus, we filtered out pages where video was the primary medium. To further ensure the quality of the extracted documents, we applied a series of heuristic filters to remove low-value content: we discarded pages that primarily consisted of a list of references, contained fewer than ten words, or featured boilerplate text used to introduce a course or section.

\subsubsection{Document Pairing}

The structured nature of the OpenWHO platform facilitated document alignment. For any given course page, the quadruplet \texttt{(language code, course id, section index, subsection index)} serves as a unique identifier. By varying the \texttt{language code}, we could accurately identify and group parallel course pages that are direct translations of one another.

After applying the quality filters described in the previous section, this pairing process yielded $2,978$ parallel documents. This set includes significant coverage for several low- and mid-resource languages, with Tetun, Albanian, Macedonian, Azerbaijani, Kazakh, Georgian, and Armenian each having more than 50 parallel documents. A breakdown of document counts per language is presented in Table~\ref{tab:docs-and-sents-per-lang}.

\subsubsection{Sentence Mining}

Traditional NMT models rely on sentence-level translation. To release a dataset that can be used for NMT evaluation, and potentially fine-tuning, we mined parallel sentences from parallel documents.

\paragraph{Annotation}
To evaluate our sentence mining pipeline, we manually annotated 10 parallel documents for 8 target low-resource languages (Macedonian, Georgian, Armenian, Azerbaijani, Tetun, Albanian, Kazakh, Tamil) that are of interest to our experiment. For each document pair, we manually segmented the source and target texts into aligned sentences (relying on back-translation for the languages we are not familiar with), ensuring one-to-one correspondence. This process yielded a reference corpus totalling $2,645$ parallel sentences. This annotated set serves as the ground truth for the subsequent sentence-splitting and alignment~evaluation.

\paragraph{Sentence splitting}
Using the target-language sentences from our manually annotated corpus as a reference, we evaluated the performance of three sentence-splitters: \texttt{NLTK} \cite{bird-loper-2004-nltk}, \texttt{Stanza} \cite{qi-etal-2020-stanza}, and \texttt{pysbd} \cite{sadvilkar-neumann-2020-pysbd}. We measured sentence splitting performance as accuracy of sentence boundaries against our ground truth segmentation. The results (shown in Appendix~\ref{sec:sentencization-perf}) indicated that \texttt{pysbd} achieves the highest accuracy overall with accuracy ranging from 82.0\% for Kazakh to 94.0\% for Tetun, but \texttt{stanza} performs better for Kazakh (89.1\%), Tetun (94.6\%) and Georgian (93.2\%). NLTK's performance was generally lower than the other two. Based on these findings, we selected the best-performing tool (either \texttt{pysbd} or \texttt{stanza}) on a per-language basis to segment the entire corpus.

\paragraph{Sentence alignment}
After sentence splitting, we aligned sentences to create parallel pairs.
Here we rely on sentence semantic similarity, using LaBSE (\textit{Language-agnostic BERT Sentence Embedding} \cite{feng-etal-2022-language}), which supports all our languages of interest except Tetun (as a consequence, for Tetun, we first translated target sentences back to English before encoding).
Because the OpenWHO documents are relatively short, this approach is highly effective: when evaluated against our manually annotated ground truth, the method yielded F1 scores ranging from 98.6\% (for Tetun) to 100\% (for Kazakh and Georgian).

\paragraph{Quality Control and Filtering}

Finally, to ensure the quality of the mined sentence pairs, we implemented an additional filtering stage based on empirical rules. We removed sentence pairs that were likely to be misaligned or uninformative for translation tasks. This included removing (1) pairs where the source English sentence contained fewer than five words, as these are often section headers or fragments; (2) pairs where the target-language side was in English; and (3) sentence pairs that were exact duplicates across different course pages, which often correspond to repeated instructions or boilerplate phrases.

Starting from an initial pool of $43,732$ candidate sentence pairs, we arrived at a final, clean set of \textbf{26,$\,$824} parallel sentences. This includes nine low-resource languages with over 200 parallel sentences (Macedonian, Georgian, Armenian, Azerbaijani, Tetun, Albanian, Kazakh, Somali, Sinhala). The count of sentence pairs per language is detailed in Table~\ref{tab:docs-and-sents-per-lang}.

\subsection{Dataset Statistics}

The resulting OpenWHO corpus comprises \textbf{2,$\,$978} parallel documents and \textbf{26,$\,$824} aligned parallel sentences between English and over 20 other languages. The corpus contains a mix of high-resource and low-resource languages, with significant depth in the latter, including six with over 1,000 parallel sentences (Macedonian, Georgian, Armenian, Albanian, Kazakh, and Tetun). A key feature of this dataset is its origin: all content is expert-authored and professionally translated, providing high-fidelity, domain-specific text that is a level above standard web-crawled corpora in terms of quality and consistency. The data is structured at both the document and sentence level, enabling experiments in document-level machine translation, terminology extraction, and domain adaptation. However, a potential weakness of this dataset is its unbalanced language distribution, as not all courses were translated into all languages.

\subsection{Data Availability}

With permission from the WHO, we release this dataset under a Creative Commons NonCommercial license (CC BY-NC 4.0), allowing re-use, modification and distribution for non-commercial use, while requiring attribution. Data will be available both at the document level and at the sentence level.

\section{Experiments}
\label{sec:experiments}

% Introduce the "context dilemma": a combination of several factors
% - translation benefits from context
% - models can trip over themselves and hallucinate more when there is more content
% - models can benefit from their past translation as context in the TGT lang, almost can be thought of as a CoT

Having established an evaluation corpus for document-level low-resource MT in the health domain, we now turn to investigating what models perform best on this dataset, and how context utilisation strategies affect LLM performance on this dataset. Our experimental design addresses a fundamental tension in document-level translation: while broader context can improve coherence and terminological consistency, it may also introduce noise or lead to error propagation.

We work with the following research questions:

\begin{itemize}
    \item \textbf{RQ1}: How do state-of-the-art LLMs compare to traditional NMT models for health low-resource translation?
    \item \textbf{RQ2}: What is the most effective context strategy for LLM-based translation into low-resource languages? (sentence-level, document-level, sliding sentence window, etc)
    \item \textbf{RQ3}: How does model capability interact with these context strategies?
\end{itemize}

% Our evaluation spans both specialised (health, literature) and general domains, using models of varying scales.

\subsection{Experimental Setup}

\paragraph{Datasets}
We evaluate on two datasets, always in the EN-XX direction. The first is our newly introduced OpenWHO corpus. To ensure a controlled comparison across languages, we focus our experiments on a single, extensively translated course: ``Infection Prevention and Control through Hand Hygiene (IPC-HH)''. We select the nine low- to mid-resource languages available for this course for our evaluation: Albanian (sqi), Armenian (hye), Azerbaijani (aze), Georgian (kat), Kazakh (kaz), Macedonian (mkd), Sinhala (sin), Somali (som), and Tetun (tet). For a comparison with high-resource languages, we separately evaluate on French (fra), Russian (rus) and Spanish (spa).

To test the generalisability of our findings beyond the health domain, we also evaluate on the WMT24++ benchmark \cite{deutsch_wmt24_2025}, an expansion of the WMT24 dataset to 55 languages. To align with our research focus, we select a sample of five low- to mid-resource languages present in this dataset: Bulgarian (bul), Serbian (srp), Swahili (swh), Tamil (tam), and Zulu (zul). Because this dataset is available at the paragraph level, for our sentence-level analysis, we split paragraphs into aligned sentences using Gemini 2.5 Flash.

\paragraph{Models}
Our model selection is designed to compare modern LLMs (both open and closed weights) against conventional NMT baselines.
For NMT baselines, we select NLLB-200 (3.3B \& 54B, \citeauthor{costa-jussa_scaling_2024}, \citeyear{costa-jussa_scaling_2024}) and MADLAD-400 10B \cite{kudugunta_madlad-400_2023}, both of which cover languages covered in our evaluation (except Tetun for NLLB). For LLMs, we select Gemini 2.5 Flash \cite{Gemini25technicalreport}, a powerful closed-weight model, DeepSeek-V3 671B \cite{deepseekai2024deepseekv3technicalreport}, which represents the state-of-the-art in open-weight models at the time of our experiments, and Gemma 3 27B \cite{gemma_2025}, a smaller LLM with broad multilingual support. We run all model calls through OpenRouter.\footnote{\href{https://openrouter.ai/}{https://openrouter.ai/}}

\begin{table}[t]
    \centering
    \begin{tabular}{l l l}
        \toprule
        & \textbf{ChrF $\uparrow$} & \textbf{MetricX $\downarrow$} \\
        \midrule
        \multicolumn{3}{l}{\textbf{OpenWHO} (9 low-res langs)} \\
        \quad NLLB 54B & 50.52 & 3.45 \\
        % For ChrF, a positive change is good (goodup). For MetricX, a negative change is good (gooddown).
        \quad Gemini & 55.32 \goodup{4.79} & 3.10 \gooddown{-0.43} \\
        \quad DeepSeek-v3 & 49.38 \baddown{-1.14} & 3.92 \badup{0.39} \\
        \quad Gemma 3 & 48.01 \baddown{-2.51} & 4.24 \badup{0.71} \\
        \midrule
        \multicolumn{3}{l}{\textbf{WMT24++ literary} (5 low-res langs)} \\
        \quad NLLB 54B & 43.00 & 5.83 \\
        \quad Gemini & 50.66 \goodup{7.66} & 3.76 \gooddown{-2.07} \\
        \quad DeepSeek-v3 & 46.88 \goodup{3.88} & 4.57 \gooddown{-1.26} \\
        \quad Gemma 3 & 44.45 \goodup{1.45} & 5.26 \gooddown{-0.57} \\
        \midrule
        \multicolumn{3}{l}{\textbf{WMT24++ news} (5 low-res langs)} \\
        \quad NLLB 54B & 53.58 & 3.45 \\
        \quad Gemini & 54.83 \goodup{1.24} & 2.69 \gooddown{-0.76} \\
        \quad DeepSeek-v3 & 51.40 \baddown{-2.18} & 3.42 \gooddown{-0.04} \\
        \quad Gemma 3 & 50.71 \baddown{-2.87} & 3.61 \badup{0.16} \\
        \bottomrule
    \end{tabular}
    \caption{Average performance per model, with score difference from NLLB 54B. Modern LLMs like Gemini outperform NLLB on specialised domain low-resource MT, like health or literary fiction. See Tables~\ref{tab:nmt-vs-llm-openwho} and \ref{tab:nmt-llm-wmt24pp-literary} for scores per language, which vary from 37 to 63 ChrF. }
    \label{tab:nmt-vs-llm-averages}
\end{table}

\paragraph{Metrics}
We primarily evaluate with ChrF++ \cite{popovic_chrf_2017}, an n-gram based metric which has been shown to correlate better with human judgement than BLEU \cite{papineni-etal-2002-bleu} particularly for morphologically rich languages like Kazakh or Georgian.
To validate results found with ChrF++, we also evaluate with MetricX-24\footnote{\href{https://huggingface.co/google/metricx-24-hybrid-large-v2p6-bfloat16}{google/metricx-24-hybrid-large-v2p6-bfloat16}} \cite{juraska_metricx-24_2024} and AutoMQM \cite{fernandes_devil_2023}.
MetricX is a neural metric which correlates better with human judgement than ChrF++ for high-resource languages. While it has not been evaluated on low-resource languages, it is based on mT5 \cite{xue-etal-2021-mt5}, which has been pretrained on all languages in our study, aside from Tetun.
AutoMQM uses a large language model to characterise translation errors using MQM \cite{burchardt-2013-multidimensional}. To avoid self-preference bias that may arise from using the same LLM for AutoMQM as that used for translation \cite{wataoka_self-preference_2025}, we run AutoMQM with Kimi K2 \cite{kimik2}.

% Anecdotally, \cite{falcao_comet_2024} find that ChrF correlate better with human judgement with COMET-22 on English-Maltese
% And \citet{avramidis-etal-2024-machine} find that ChrF performs better than Gemba (gpt4-based) on en-de!

\paragraph{Translation Strategies}
For LLM translation, we rely on a fixed one-shot prompt (Appendix~\ref{sec:appendix-prompts}), and we systematically evaluate five translation strategies that introduce contextual information in different ways. Detailed in Table~\ref{tab:translation_strategies}, these include translating sentences one at a time, translating sentences with some surrounding context, and translating entire documents at once.
For NMT models (NLLB and MADLAD), as they were trained at the sentence level, we evaluate only at the sentence~level.
To ensure a fair comparison across models and strategies, all outputs, including those generated at the document level, are segmented and evaluated at the sentence level against the reference translations.

\subsection{Results}
\label{sec:exp-results}

\paragraph{LLMs outperform NMT on health low-resource translation (RQ1).}
On OpenWHO, Gemini 2.5 Flash, when translating at the document level, outperforms NLLB 54B across all languages,\footnote{We exclude Tetun from this comparison, as NLLB does not support it.} by an average of +4.79 ChrF points (Table~\ref{tab:nmt-vs-llm-averages}). MetricX and AutoMQM results confirm this overall trend. However, other LLMs evaluated (DeepSeek-v3 and Gemma~3) are still outperformed by NLLB-54B, albeit by a small margin for DeepSeek. This means that among open weight models, NLLB-54B is still the preferred choice. Further, at equivalent performance before fine-tuning, LLMs require far more computation, with around one order of magnitude more parameters for the same performance (e.g. DeepSeek-v3 671B roughly equivalent to NLLB 54B; Gemma~3 27B equivalent to NLLB 3.3B).

\paragraph{Error analysis: Gemini vs NLLB} Error analysis using AutoMQM (Table~\ref{tab:automqm-openwho}) shows that Gemini translations contain substantially fewer critical errors than NLLB, with less mistranslations (where target text does not accurately represent the source meaning) and less incorrect terminology, at the cost however of more omissions (where target text is missing information present in the source) and overtranslations (target text more specific than the~source). 

\paragraph{On high-resource languages, NLLB and LLMs are very close to each other.}
On our sample of high-resource OpenWHO languages (French, Russian, Spanish), the average scores for NLLB 54B, Gemini, DeepSeek, and Gemma 3 are all remarkably close to each other, as measured by both ChrF (averages in the 59-62 range for all 4 models) and MetricX (averages in the 2.3-2.4 range). This result indicates that in the health domain, the advantage of LLMs over NMT is more pronounced on low-resource languages compared to high-resource. Unsurprisingly, performance on high-resource languages is notably higher than on low-resource ones, with a gap of 7-12 ChrF points between high-resource and low-resource across all~models.

\begin{table}[!t]
    \centering
    \begin{tabular}{l r r}
        \toprule
        \textbf{Doc vs sent} & \textbf{ChrF $\Delta$} & \textbf{MetricX $\Delta$} \\
        \midrule
        \multicolumn{3}{l}{\textbf{OpenWHO} (9 low-res langs)} \\
        \quad Gemini & \goodup{3.62} & \neutral{0.00} \\
        \quad DeepSeek & \goodup{2.00} & \neutral{0.02} \\
        \quad Gemma3 & \baddown{-0.21} & \badup{0.24} \\
        \midrule
        \multicolumn{3}{l}{\textbf{WMT literary} (5 low-res langs)} \\
        \quad Gemini & \goodup{6.37} & \gooddown{-1.18} \\
        \quad DeepSeek & \goodup{3.34} & \gooddown{-0.79} \\
        \quad Gemma3 & \goodup{2.06} & \gooddown{-0.14} \\
        \midrule
        \multicolumn{3}{l}{\textbf{WMT news} (5 low-res langs)} \\
        \quad Gemini & \goodup{1.24} & \gooddown{-0.08} \\
        \quad DeepSeek & \baddown{-0.82} & \gooddown{-0.11} \\
        \quad Gemma3 & \baddown{-0.14} & \badup{0.13} \\
        \bottomrule
    \end{tabular}
    \caption{Performance difference for document-level vs sentence-level translation, averaged across languages. In specialised domains (health, literary fiction), the larger the LLM, the more it benefits from doc-level translation. See Tables~\ref{tab:llm-strategy-openwho} and \ref{tab:llm-strategy-wmt} for scores per language.}
    \label{tab:llm-doc-vs-sent-avg}
\end{table}

\paragraph{LLMs tend to work best at the document level, for specialised domains (RQ2).}
On OpenWHO, both Gemini and DeepSeek translate best at the document level, with +3.62 and +2.00 ChrF points over sentence-level translation respectively (Table~\ref{tab:llm-doc-vs-sent-avg}), but no measurable improvements in MetricX scores. On WMT24++ literary, the advantage of document-level over sentence-level is even clearer, with +6.37 Chrf points for Gemini, +3.34 for DeepSeek, and similar improvements in MetricX scores (Table~\ref{tab:llm-strategy-wmt}). For Gemma 3 27B however, additional context from document-level only marginally improves translation accuracy, on both OpenWHO and WMT24++ literary. Overall, we observe a trend where \textbf{the larger the LLM, the more it benefits from document-level translation (RQ3)} over sentence-level translation.

\paragraph{In the general domain, the advantage of modern LLMs and document-level translation are less clear.}
On the WMT24++ news set, we do not see meaningful accuracy improvements for document-level over sentence-level translation, using either metric (ChrF and MetricX). We also see less variation in scores between models on this domain, both when comparing NLLB to LLMs, and when comparing LLMs with each other. Overall, the advantage of document-level translation over sentence-level translation for low-resource MT is not uniform across domains and models.

% \paragraph{In the general domain, the benefit of document-level translation is less clear.} Across LLMs, we do not see large accuracy improvements on the WMT24++ news set, for document-level versus sentence-level translation, with +1.25 ChrF for Gemini, -0.82 for DeepSeek, and -0.14 for Gemma~3.

% \paragraph{Sentence + context tends to not help translation accuracy.}
% Across all LLMs, translation strategies that rely on individual sentence translation with additional source context, such as ``sentence window'' and ``sentence + doc context'' strategies in Table~\ref{tab:llm-strategy-openwho}, do not tend to result in translation accuracy improvements over pure sentence-level translation, despite their additional computing cost. This suggests that \textbf{context helps mostly with target-level consistency} (maintaining terminology coherence in the generated translation) \textbf{rather than with source-level contextualisation} for understanding ambiguous or context-dependent source text.

\section{Discussion}

% to mention here:
% - maybe LLMs benefit from extended pretraining, which means better handling of domains that are poorly represented in low-res corpora 
% - Gemini beats NLLB 54B thanks to its ability to get document-level context. When translating at the sentence-level, for all 3 datasets evaluated (OpenWHO, WMT24++ literary, WMT24++ news), Gemini is within 1.5 ChrF point of NLLB 54B. It's only when translating at the document-level that it differentiates itself.
% - document-level vs domain: some domains require context to select terminology (e.g. tone for literature, specialised terminology for health), maybe news has more self-contained sentences
% - Recommendations: Evaluate LLMs at the document-level; evaluate very large LLMs to take full advantage of document context; evaluate per-language, as average model ranking is not a good reflection of per-language performance.

% principal findings
Our experiments present \textbf{three key findings}: First, modern LLMs tend to outperform NMT (e.g. Gemini outperforms NLLB 54B) on low-resource translation in specialised domains (health with OpenWHO, literary text with WMT24++). Second, modern LLMs translate best at the document-level in specialised domains (health and literary), but the advantage of document-level translation is less clear for smaller models and for the general domain. Third, other context-utilisation strategies (e.g. sentence window, document context with one sentence at a time) tend to perform less well than whole-document translation.

\paragraph{Why Gemini outperforms NLLB in low-resource specialised domain MT (RQ1)}
Our investigation into performance differences between LLMs and NMT models reveals that Gemini's advantage over NLLB 54B in specialised domains stems directly from its ability to leverage document-level context. When both models are constrained to sentence-level translation, their performance is very similar across all three datasets evaluated (OpenWHO, WMT24++ literary, and WMT24++ news, all within a narrow 1.5 ChrF point margin). It is only when Gemini is provided with the full document that it establishes a clear performance lead.

\paragraph{The role of context strategy across domains}
% some domains require context to select terminology (e.g. tone for literature, specialised terminology for health). News articles, which may be designed for skimming and for sharing, do not require context as much.
% Maybe larger models 
Our findings show that the optimal context strategy depends on both text domain (RQ2) and model capability (RQ3). The benefit of document-level translation is most pronounced in specialised domains like health and literature, potentially because their discourse structure requires a high degree of linguistic coherence for both accuracy (e.g. correct health terminology) and stylistic integrity (e.g. sustained narrative tone). In contrast, the news domain may rely more on self-contained sentences that allow skimming and quoting, reducing the benefit of context.
Further, our results indicate that smaller models only gain marginal benefits from document context, potentially lacking the capacity to maintain coherence without introducing noise.

Our findings, particularly the dependence of context utility on model capability and domain specificity, offer a nuanced picture for where document-level context is most useful, which may explain past work that either did not \cite{li-etal-2020-multi-encoder,appicharla-etal-2024-case,koneru_contextual_2024} or did \cite{wang_document-level_2023,post_escaping_2024,wu_adapting_2024} find added benefits from contextual level~translation.

\paragraph{Recommendations}
Based on our results, we offer three recommendations for researchers working on low-resource MT: 
\begin{enumerate}
    \item \textbf{Evaluate LLMs at the document level for specialised domains.} Sentence-level evaluation can mask the advantage of modern LLMs, which lies in their ability to use context.
    \item \textbf{Utilise the most capable LLMs to maximise the benefit of document context.} The performance gains from document-level translation are most significant with the largest models.
    \item \textbf{Analyse performance on a per-language basis.} Average model rankings do not always reflect performance on individual languages, making granular analysis essential for model~selection.
\end{enumerate}

\paragraph{Future directions}
Several avenues for future work emerge from our findings. First, the development of reliable evaluation metrics tailored to low-resource MT in the health domain. Second, further exploration of strategies to optimise LLM-based translation for low-resource health contexts, such as fine-tuning on domain-specific data or different prompting techniques. Third, the creation of evaluation benchmarks for low-resource health on other tasks, such as question answering, which OpenWHO could be leveraged for.

\section{Conclusion}

In this work, we introduced OpenWHO, a high-quality parallel corpus for health MT, with a focus on low-resource languages. Sourced from the World Health Organization’s expert-authored materials, it addresses a gap in evaluation resources and provides a benchmark for future research at the intersection of health and low-resource languages. The dataset strengths include (1) the grounding of its source English text in evidence‐based WHO guidance (2)~its professional translation into various languages and (3) its availability at both the document and sentence level. However, OpenWHO is language imbalanced (not all courses were translated into all languages), which can limit its comparative value.

Our experiments demonstrate that modern LLMs, when provided with full document-level context, outperform traditional NMT models on low-resource translation in specialised domains like health and literature. We found that this advantage is most pronounced for the largest models and diminishes in the general (news) domain, highlighting that the utility of context depends on both model capability and domain complexity. Our work underscores the potential of document-aware LLMs to improve translation quality in high-impact settings, while also revealing the critical need for domain-specific evaluation benchmarks and context-aware translation strategies.

\section{Limitations}

\paragraph{Metrics}
Our findings rely exclusively on automated metrics (ChrF, MetricX, AutoMQM). While these metrics give a useful signal when they all agree, we have limited ability to resolve differences when they arise. ChrF is a recognised standard for low-resource MT but may not always correlate well with human judgement \cite{wang-etal-2024-afrimte}; MetricX and AutoMQM have not been evaluated on low-resource languages, let alone in the health domain. Overall, more work is needed to determine what is the right metric for low-resource health MT, including a comprehensive human evaluation to validate our findings and gain a more nuanced understanding of translation~quality.

\paragraph{Generalisability across other domains}
In our experiments on context utilisation, we rely on two specialised domains: health (OpenWHO) and literary fiction (WMT24++). While we find similar trends, our findings may not generalise to other specialised domains, such as legal, financial, or technical texts. The specific characteristics of each domain may influence the utility of document-level context, and a broader, structured evaluation across multiple domains would be needed to draw more general conclusions.

\paragraph{Caveats of a direct comparison between LLMs and NMT}
While document-level LLM translation beats sentence-level NMT translation for the languages and specialised domains we evaluate on, this comparison might be unfair to NMT models, which could be adapted to benefit from document-level context for a more equivalent comparison, and have far fewer model parameters at equivalent performance levels. In practice, LLM outputs could be leveraged for knowledge distillation, creating smaller, domain-specific models that retain much of the performance advantage while being more efficient \cite{gibert_scaling_2025}.
% Therefore, while LLMs may be the best choice when maximal accuracy is the sole goal, distilled NMT models might still offer a better balance of performance and computational cost for certain applications.

\paragraph{Dataset language imbalance}
Finally, the OpenWHO dataset itself has limitations. Its language distribution is imbalanced, as not all source materials were translated into every target language. This can constrain its utility for direct cross-language~comparisons.

\section{Ethics Statement}

\paragraph{Consent}
This work adheres to ethical guidelines for data collection and research in natural language processing. The OpenWHO corpus was compiled from the WHO's e-learning platform with explicit authorization from the WHO for both data collection and public release. Our work aligns with the WHO's mission to disseminate health information globally and respects their ownership of the~content.

\paragraph{Dual use and societal impact}
We have carefully considered the potential for dual use of the OpenWHO dataset and our research findings. Our primary objective is to enhance access to health education material by improving MT for low-resource languages in the high-stakes health domain. The dataset comprises expert-authored, professionally translated public health materials, limiting risks of misuse. The humanitarian and public health benefits of facilitating information access in underserved languages significantly outweigh dual-use~concerns.

\section*{Acknowledgements}

We are deeply grateful to the World Health Organization (WHO) for their collaboration and for granting us permission to collect and publicly release the OpenWHO dataset. In particular, we would like to express our sincere gratitude to Heini Utunen, Corentin Piroux, and Melissa Attias for their support and guidance on this project.

This research was supported by The University of Melbourne’s Research Computing Services and the Petascale Campus Initiative.

\bibliography{custom, anthology_0, anthology_1}

\appendix

% 5-shot Gemini (sent-level) TET 51.64
% 5-shot Gemini (doc-level) 52.39
% SO: 5-shot actually makes the quality lower, not higher

\clearpage

\section{Documents and sentences per language}

\begin{minipage}{\textwidth}
\centering
\small
\begin{tabular}{llrr}
\toprule
\textbf{Language} & \textbf{Script} & \textbf{Number of documents} & \textbf{Number of sentences} \\
\midrule
Russian (rus)        & Cyrillic & 315 & 2194 \\
French (fra)         & Latin & 301 & 2385 \\
Arabic (ara)         & Arabic & 293 & 2623 \\
\textbf{Macedonian (mkd)}  & Cyrillic & 254 & 3695 \\
Ukrainian (ukr)      & Cyrillic & 204 & 1632 \\
Chinese (zho)        & Chinese & 203 & 871 \\
Spanish (spa)        & Latin & 149 & 596 \\
\textbf{Georgian (kat)}    & Georgian & 131 & 2151 \\
\textbf{Armenian (hye)}    & Armenian & 125 & 1982 \\
\textbf{Kazakh (kaz)}      & Cyrillic & 103 & 936 \\
\textbf{Azerbaijani (aze)} & Latin & 98  & 1677 \\
Turkish (tur)        & Latin & 81  & 1093 \\
Indonesian (ind)     & Latin & 80  & 329 \\
Dutch (nld)          & Latin & 77  & 1082 \\
\textbf{Albanian (sqi)}    & Latin & 74  & 1029 \\
\textbf{Tetun (tet)}       & Latin & 67  & 1086 \\
Portuguese (por)     & Latin & 62  & 279 \\
Hindi (hin)          & Devanagari & 28  & 35 \\
Tamil (tam)          & Tamil & 26  & 207 \\
\textbf{Sinhala (sin)}     & Sinhala & 25  & 214 \\
Persian (fas)        & Perso-Arabic & 25  & 56 \\
\textbf{Amharic (amh)}     & Ethiopic & 22  & 25 \\
\textbf{Marathi (mar)}     & Devanagari & 21  & 19 \\
\textbf{Somali (som)}      & Latin & 20  & 224 \\
Italian (ita)        & Latin & 20  & 60 \\
\textbf{Lao (lao)}         & Lao & 18  & 29 \\
\textbf{Yoruba (yor)}      & Latin & 17  & 14 \\
\textbf{Burmese (mya)}     & Burmese & 15  & 32 \\
\textbf{Swahili (swa)}     & Latin & 14  & 107 \\
Vietnamese (vie)     & Latin & 11  & 13 \\
Catalan (cat)        & Latin & 10  & 8 \\
\textbf{Pushto (pus)}      & Perso-Arabic & 8   & 22 \\
\textbf{Hausa (hau)}       & Latin & 8   & 28 \\
Thai (tha)           & Thai & 7   & 8 \\
\textbf{Shan (shn)}        & Shan & 7   & 3 \\
\textbf{S'gaw Karen (ksw)} & Karen & 7   & 2 \\
Japanese (jpn)       & Japanese & 6   & 9 \\
Bulgarian (bul)      & Cyrillic & 6   & 9 \\
\textbf{Bengali (ben)}     & Bengali & 6   & 8 \\
\textbf{Urdu (urd)}        & Perso-Arabic & 4   & 10 \\
\textbf{Telugu (tel)}      & Telugu & 4   & 4 \\
Greek (ell)          & Greek & 4   & 4 \\
Serbian (srp)        & Latin & 3   & 4 \\
Polish (pol)         & Latin & 3   & 5 \\
\textbf{Panjabi (pan)}     & Gurmukhi & 3   & 4 \\
\textbf{Oriya (ori)}       & Odia & 3   & 4 \\
\textbf{Kurdish (kur)}     & Latin/Arabic & 3   & 3 \\
\textbf{Tajik (tgk)}       & Cyrillic & 2   & 1 \\
Romanian (ron)       & Latin & 2   & 6 \\
\textbf{Nigerian Pidgin (pcm)} & Latin & 2 & -- \\
\textbf{Lingala (lin)}     & Latin & 1   & 7 \\
\bottomrule
\end{tabular}
\captionof{table}{Number of OpenWHO documents and sentences per language. Low-resource languages are in \textbf{bold}.}
\label{tab:docs-and-sents-per-lang}
\end{minipage}

\clearpage

\section{Sentence splitting performance}
\label{sec:sentencization-perf}

\begin{minipage}{\textwidth}
\small
\begin{tabular}{lcccccccc}
\toprule
\textbf{Method} & \textbf{Tamil} & \textbf{Armenian} & \textbf{Azerbaijani} & \textbf{Macedonian} & \textbf{Kazakh} & \textbf{Tetun} & \textbf{Georgian} & \textbf{Albanian} \\
\midrule
pysbd           & \textbf{86.8}  & \textbf{87.8}     & \textbf{90.9}        & \textbf{89.6}       & 82.0            & 94.0           & 92.1           & \textbf{91.3}     \\
nltk            & 80.7           & 35.6              & 85.6                 & 82.9                & 82.0            & 85.5           & 91.8           & 88.0              \\
stanza          & 80.3           & 83.4              & 88.9                 & 84.0                & \textbf{89.1}   & \textbf{94.6}  & \textbf{93.2}  & 76.1              \\
\bottomrule
\end{tabular}
\captionof{table}{Sentence splitting performance (Accuracy \%) per language. The best score for each language is highlighted in bold.}
\label{tab:sentencizer-acc-f1}
\end{minipage}

\clearpage

\section{OpenWHO performance per language}

% NMT vs LLM table
\begin{minipage}{\textwidth}
\centering
\small
\begin{tabularx}{\textwidth}{l *{10}{>{\centering\arraybackslash}X}}
\toprule
\textbf{Model} & \textbf{mkd} & \textbf{kaz} & \textbf{kat} & \textbf{hye} & \textbf{aze} & \textbf{sqi} & \textbf{tet} & \textbf{som} & \textbf{sin} & \textbf{AVG} \\
\midrule
% --- NMT Baselines ---
MADLAD-400 10B & 58.37 / 3.25 & 47.29 / 4.59 & 15.81 / 14.09 & 37.27 / 5.25 & 40.54 / 6.24 & 54.97 / 3.87 & 44.29 / 7.35 & 48.13 / 6.44 & 39.48 / 6.04 & 42.73 / 6.22 \\
NLLB-200 3.3B & 50.39 / 3.18 & 42.94 / 3.64 & 38.27 / 4.22 & 39.19 / 4.20 & 45.23 / 4.32 & 57.50 / 3.09 & -- / -- & 47.05 / 4.40 & 39.69 / 3.40 & 45.03 / 3.81 \\
NLLB-200 54B & 56.17 / 2.94 & 56.55 / \textbf{3.15} & 43.90 / 4.21 & 42.91 / 3.62 & 48.78 / 3.92 & 59.01 / 2.84 & -- / -- & 48.23 / 4.50 & 48.64 / 3.04 & 50.52 / 3.53 \\
\midrule
% --- LLM Systems (reporting best performance) ---
Gemma-3 27B & 58.52 / 2.95 & 48.90 / 3.99 & 43.76 / 4.66 & 43.37 / 4.09 & 46.09 / 4.36 & 58.12 / 3.11 & 36.85 / 8.82 & 46.01 / 5.62 & 39.32 / 5.11 & 48.01 / 4.24 \\
DeepSeek-V3 671B & 59.07 / 2.98 & 50.39 / 3.69 & 46.27 / 3.78 & 47.41 / 3.39 & 47.84 / 3.73 & 59.02 / 2.89 & 47.64 / 7.12 & 43.36 / 6.18 & 41.70 / 4.68 & 49.38 / 3.92 \\
Gemini 2.5 Flash & \textbf{62.83} / \textbf{2.65} & \textbf{57.41} / 3.17 & \textbf{50.28} / \textbf{3.00} & \textbf{49.40} / \textbf{3.00} & \textbf{52.62} / \textbf{3.51} & \textbf{60.33} / \textbf{2.72} & \textbf{51.86} / \textbf{6.22} & \textbf{55.09} / \textbf{4.15} & \textbf{54.58} / \textbf{2.58} & \textbf{55.32} / \textbf{3.10} \\
\bottomrule
\end{tabularx}
\captionof{table}{Overall performance (ChrF++ / MetricX) on the OpenWHO test set. LLM scores represent their optimal strategy (max in Table~\ref{tab:llm-strategy-openwho}). The best score in each column is in \textbf{bold}.}
\label{tab:nmt-vs-llm-openwho}
\end{minipage}

\vspace{2em}
\begin{minipage}{\textwidth}
\centering
\small
\begin{tabularx}{\textwidth}{ll *{10}{>{\centering\arraybackslash}X} >{\centering\arraybackslash}X}
\toprule
\textbf{Model} & \textbf{Strategy} & \textbf{mkd} & \textbf{kaz} & \textbf{kat} & \textbf{hye} & \textbf{aze} & \textbf{sqi} & \textbf{tet} & \textbf{som} & \textbf{sin} & \textbf{AVG} & \textbf{$\Delta$} \\
\midrule
\multirow{5}{*}{\textbf{Gemini 2.5 Flash}}
 & Sentence level           & \num{57.79} & \num{54.86} & \num{46.24} & \num{46.13} & \num{48.62} & \num{57.02} & \num{46.83} & \num{52.24} & \num{50.89} & \num{51.18} & -- \\
 & Sentence window          & \num{58.53} & \num{54.81} & \num{47.20} & \num{45.60} & \num{48.86} & \num{58.47} & \num{46.68} & \num{48.91} & \num{50.11} & \num{51.02} & \num{-0.16} \\
 & Sentence + doc context   & \num{58.64} & \num{54.85} & \num{45.35} & \num{46.44} & \num{50.19} & \num{57.84} & \num{47.27} & \num{51.30} & \num{51.43} & \num{51.48} & \num{+0.30} \\
 & Document level           & \num{62.53} & \textbf{\num{57.41}} & \num{49.01} & \num{49.06} & \num{51.58} & \num{59.95} & \num{51.51} & \num{55.04} & \textbf{\num{54.58}} & \num{54.52} & \num{+3.34} \\
 & Doc-level + self-correct & \textbf{\num{62.83}} & \num{56.30} & \textbf{\num{50.28}} & \textbf{\num{49.40}} & \textbf{\num{52.62}} & \textbf{\num{60.33}} & \textbf{\num{51.86}} & \textbf{\num{55.09}} & \num{54.47} & \textbf{\num{54.80}} & \num{+3.62} \\
\midrule
\multirow{5}{*}{\textbf{DeepSeek-V3}}
 & Sentence level           & \num{56.99} & \num{49.15} & \num{42.61} & \num{44.29} & \num{45.24} & \num{56.97} & \num{43.53} & \num{41.18} & \num{40.71} & \num{46.74} & -- \\
 & Sentence window          & \num{57.81} & \num{49.54} & \num{44.88} & \num{45.40} & \num{47.64} & \num{58.07} & \num{44.48} & \num{42.78} & \textbf{\num{41.70}} & \num{48.03} & \num{+1.29} \\
 & Sentence + doc context   & \num{58.31} & \num{49.65} & \num{44.16} & \num{44.58} & \num{46.64} & \num{57.90} & \num{43.30} & \num{42.11} & \num{40.22} & \num{47.43} & \num{+0.69} \\
 & Document level           & \textbf{\num{59.07}} & \textbf{\num{50.39}} & \textbf{\num{46.27}} & \num{44.95} & \textbf{\num{47.84}} & \textbf{\num{59.02}} & \textbf{\num{47.64}} & \num{42.85} & \num{40.67} & \textbf{\num{48.74}} & \num{+2.00} \\
 & Doc-level + self-correct & \num{55.70} & \num{46.27} & \num{43.27} & \textbf{\num{47.41}} & \num{47.41} & \num{57.10} & \num{45.65} & \textbf{\num{43.36}} & \num{41.30} & \num{47.50} & \num{+0.76} \\
\midrule
\multirow{5}{*}{\textbf{Gemma-3 27B}}
 & Sentence level           & \num{58.14} & \num{48.29} & \textbf{\num{43.76}} & \textbf{\num{43.37}} & \num{45.83} & \textbf{\num{58.12}} & \num{35.27} & \textbf{\num{46.01}} & \num{38.56} & \textbf{\num{46.37}} & -- \\
 & Sentence window          & \num{58.33} & \num{48.53} & \num{43.09} & \num{43.01} & \num{45.73} & \num{57.97} & \num{32.69} & \num{45.35} & \num{37.70} & \num{45.82} & \num{-0.55} \\
 & Sentence + doc context   & \num{56.64} & \num{47.52} & \num{41.02} & \num{40.22} & \num{43.05} & \num{56.79} & \num{32.62} & \num{44.44} & \num{38.01} & \num{44.48} & \num{-1.89} \\
 & Document level           & \textbf{\num{58.52}} & \textbf{\num{48.90}} & \num{40.92} & \num{42.48} & \textbf{\num{46.09}} & \num{57.51} & \num{36.52} & \num{45.23} & \textbf{\num{39.32}} & \num{46.17} & \num{-0.21} \\
 & Doc-level + self-correct & \num{57.40} & \num{46.17} & \num{42.63} & \num{41.49} & \num{46.01} & \num{57.31} & \textbf{\num{36.85}} & \num{45.39} & \num{38.33} & \num{45.73} & \num{-0.64} \\
\bottomrule
\end{tabularx}
\captionof{table}{Effect of different context strategies on LLM performance on the OpenWHO test set (ChrF++). The `$\Delta$` column shows the change relative to the `sentence level` baseline.}
\label{tab:llm-strategy-openwho}
\end{minipage}

\clearpage

\section{WMT24++ performance per language}

% NMT vs LLM, news
\begin{minipage}{\textwidth}
\centering
\small
\begin{tabularx}{\textwidth}{l *{6}{>{\centering\arraybackslash}X}}
\toprule
\textbf{Model} & \textbf{tam} & \textbf{zul} & \textbf{bul} & \textbf{srp} & \textbf{swh} & \textbf{AVG} \\
\midrule
% --- NMT Baselines ---
NLLB-200 3.3B     & 43.85 / 3.52    & \textbf{63.35} / \textbf{3.17} & 58.39 / 3.09          & 51.59 / 3.16          & 52.17 / 4.62          & 53.87 / 3.51 \\
NLLB-200 54B       & 45.52 / 3.38         & 58.39 / 3.23         & 59.80 / 2.78          & 53.18 / 2.8          & 51.02 / 5.07          & 53.58 / 3.45 \\
MADLAD-400 10B     & 40.68 / 3.98 & 38.24 / 5.69 & 59.40 / 2.88 & 47.38 / 5.38 & 46.02 / 7.1 & 46.34 / 5.01 \\
\midrule
% --- LLMs (reporting best performance) ---
Gemma-3 27B & 45.14 / 2.99 & 43.95 / 5.93 & 59.55 / 2.50 & 52.36 / 2.62 & 52.56 / 3.83 & 50.71 / 3.61 \\
DeepSeek-V3 671B & 43.95 / 3.50 & 49.41 / 4.55 & 58.27 / 2.66 & 52.53 / 2.60 & 52.84 / 3.72 & 51.40 / 3.42 \\
Gemini 2.5 Flash & \textbf{45.84} / \textbf{2.61} & 54.77 / 3.25 & \textbf{61.40} / \textbf{2.33} & \textbf{56.83} / \textbf{2.29} & \textbf{55.29} / \textbf{2.99} & \textbf{54.83} / \textbf{2.69} \\
\bottomrule
\end{tabularx}
\captionof{table}{Overall performance (ChrF++ / MetricX) on the \textbf{WMT24++ news} test set. LLM scores represent their optimal context strategy (see Table~\ref{tab:llm-strategy-wmt}).}
\label{tab:nmt-llm-wmt24pp-news}
\end{minipage}

% NMT vs LLM, literary
\vspace{2em}
\begin{minipage}{\textwidth}
\centering
\small
\begin{tabularx}{\textwidth}{l *{6}{>{\centering\arraybackslash}X}}
\toprule
\textbf{Model} & \textbf{tam} & \textbf{zul} & \textbf{bul} & \textbf{srp} & \textbf{swh} & \textbf{AVG} \\
\midrule
% --- NMT Baselines ---
NLLB-200 3.3B     & 30.74 / 7.85          & 46.51 / 4.99          & 47.29 / 4.80          & 42.17 / 5.44          & 45.16 / 6.29          & 42.37 / 5.87 \\
NLLB-200 54B       & 30.91 / 7.99          & 46.55 / 4.92          & 48.63 / 4.57          & 44.27 / 4.93          & 44.64 / 6.74          & 43.00 / 5.83 \\
MADLAD-400 10B     & 27.44 / 9.21          & 35.42 / 6.76          & 47.38 / 4.75          & 37.73 / 7.02          & 38.67 / 7.94          & 37.33 / 7.14 \\
\midrule
% --- LLMs (reporting best performance) ---
Gemma-3 27B    & 38.29 / 4.59          & 38.26 / 6.71          & 54.98 / 3.45          & 47.94 / 3.89          & 47.93 / 5.24          & 45.48 / 5.26 \\
DeepSeek-V3 671B & 38.10 / 4.99          & 44.33 / 5.64          & 53.70 / 3.55          & 49.58 / 3.61          & 48.69 / 5.08          & 46.88 / 4.57 \\
Gemini 2.5 Flash & \textbf{39.47} / \textbf{3.94} & \textbf{50.99} / \textbf{4.30} & \textbf{57.60} / \textbf{3.30} & \textbf{53.21} / \textbf{3.18} & \textbf{52.03} / \textbf{4.09} & \textbf{50.66} / \textbf{3.76} \\
\bottomrule
\end{tabularx}
\captionof{table}{Overall performance (ChrF++ / MetricX) on the \textbf{WMT24++ literary} test set. LLM scores represent their optimal context strategy (see Table~\ref{tab:llm-strategy-wmt}).}
\label{tab:nmt-llm-wmt24pp-literary}
\end{minipage}

% context strategies table - WMT24++ (Domain Split)

\vspace{2em}
\begin{minipage}{\textwidth}
\centering
\small
\begin{tabularx}{\textwidth}{ll *{6}{>{\centering\arraybackslash}X}}
\toprule
\textbf{Model} & \textbf{Strategy} & \textbf{tam} & \textbf{zul} & \textbf{bul} & \textbf{srp} & \textbf{swh} & \textbf{AVG} \\
\midrule
\multirow{6}{*}{\textbf{Gemini}} 
 & \multicolumn{7}{l}{\textit{News}} \\
 \cmidrule(l){2-8}
 & \phantom{a}Sent-level           & \textbf{45.92} / 2.65 & 53.07 / 3.33          & 59.93 / 2.47          & 53.09 / 2.39          & \textbf{55.90} / 3.03 & 53.58 / 2.77 \\
 & \phantom{a}Doc-level           & 45.84 / \textbf{2.61} & \textbf{54.77} / \textbf{3.25} & \textbf{61.40} / \textbf{2.33} & \textbf{56.83} / \textbf{2.29} & 55.29 / \textbf{2.99}          & \textbf{54.83} / \textbf{2.69}  \\
 \addlinespace[3pt]
 & \multicolumn{7}{l}{\textit{Literary}} \\
 \cmidrule(l){2-8}
 & \phantom{a}Sent-level           & 34.41 / 5.71          & 44.34 / 5.12          & 49.80 / 4.41          & 44.91 / 4.66          & 48.00 / 4.82          & 44.29 / 4.94 \\
 & \phantom{a}Doc-level           & \textbf{39.47} / \textbf{3.94} & \textbf{50.99} / \textbf{4.30} & \textbf{57.60} / \textbf{3.30} & \textbf{53.21} / \textbf{3.18} & \textbf{52.03} / \textbf{4.09} & \textbf{50.66} / \textbf{3.76} \\
\midrule
\multirow{6}{*}{\textbf{DeepSeek-V3}}
 & \multicolumn{7}{l}{\textit{News}} \\
 \cmidrule(l){2-8}
 & \phantom{a}Sent-level           & \textbf{43.95} / \textbf{3.50} & \textbf{49.41} / 4.60 & \textbf{58.27} / 2.80 & \textbf{52.53} / 2.73 & \textbf{52.84} / 3.98 & \textbf{51.40} / 3.52 \\
 & \phantom{a}Doc-level           & 43.40 / 3.55          & 48.16 / \textbf{4.55}          & 56.96 / \textbf{2.66}          & 52.23 / \textbf{2.60}          & 52.14 / \textbf{3.72}          & 50.58 / \textbf{3.42} \\
 \addlinespace[3pt]
 & \multicolumn{7}{l}{\textit{Literary}} \\
 \cmidrule(l){2-8}
 & \phantom{a}Sent-level           & 33.36 / 6.37          & 41.75 / 5.84          & 49.53 / 4.40          & 46.08 / 4.75          & 46.97 / 5.44          & 43.54 / 5.36 \\
 & \phantom{a}Doc-level           & \textbf{38.10} / \textbf{4.99} & \textbf{44.33} / \textbf{5.64} & \textbf{53.70} / \textbf{3.55} & \textbf{49.58} / \textbf{3.61} & \textbf{48.69} / \textbf{5.08} & \textbf{46.88} / \textbf{4.57} \\
\midrule
\multirow{6}{*}{\textbf{Gemma-3 27B}}
 & \multicolumn{7}{l}{\textit{News}} \\
 \cmidrule(l){2-8}
 & \phantom{a}Sent-level           & \textbf{45.14} / \textbf{2.99} & \textbf{43.95} / \textbf{5.93} & 59.55 / 2.62          & 52.36 / 2.69          & \textbf{52.56} / \textbf{3.83} & \textbf{50.71} / \textbf{3.61} \\
 & \phantom{a}Doc-level           & 45.01 / 3.17          & 42.90 / 6.47          & \textbf{60.28} / \textbf{2.50} & \textbf{52.43} / \textbf{2.62} & 52.24 / 3.94          & 50.57 / 3.74 \\
 \addlinespace[3pt]
 & \multicolumn{7}{l}{\textit{Literary}} \\
 \cmidrule(l){2-8}
 & \phantom{a}Sent-level           & 34.53 / 5.84          & \textbf{38.26} / \textbf{6.71} & 50.29 / 4.27          & 42.66 / 4.84          & 46.21 / 5.33          & 42.39 / 5.40 \\
 & \phantom{a}Doc-level           & \textbf{38.29} / \textbf{4.59} & 33.11 / 9.11          & \textbf{54.98} / \textbf{3.45} & \textbf{47.94} / \textbf{3.89} & \textbf{47.93} / \textbf{5.24} & \textbf{44.45} / \textbf{5.26} \\
\bottomrule
\end{tabularx}
\captionof{table}{Comparison of sentence-level and document-level strategies on the WMT24++ test set (ChrF++ / MetricX). The colored delta in the `AVG` column shows the change relative to the `sentence-level` baseline within each domain.}
\label{tab:llm-strategy-wmt}
\end{minipage}

\clearpage

\section{AutoMQM results}

\begin{minipage}{\textwidth}
\small
\begin{tabularx}{\textwidth}{l *{8}{>{\centering\arraybackslash}X} >{\centering\arraybackslash}X}
\toprule
\textbf{Model} & \textbf{mkd} & \textbf{kaz} & \textbf{kat} & \textbf{hye} & \textbf{aze} & \textbf{sqi} & \textbf{som} & \textbf{sin} & \textbf{AVG} \\
\midrule
\multicolumn{10}{l}{\textbf{AutoMQM score (lower is better)}} \\
\midrule
NLLB-54B & -4.72 & -3.35 & -5.44 & -4.34 & -4.27 & -3.11 & -6.08 & -4.54 & -4.48 \\
Gemini 2.5 Flash & \textbf{-2.80} & \textbf{-3.15} & \textbf{-3.12} & \textbf{-2.55} & \textbf{-3.01} & \textbf{-2.59} & \textbf{-4.88} & \textbf{-2.40} & \textbf{-3.06} \\
\midrule
\multicolumn{10}{l}{\textbf{Difference in error counts (Gemini - NLLB)}} \\
\midrule
\textbf{Error category} & \textbf{mkd} & \textbf{kaz} & \textbf{kat} & \textbf{hye} & \textbf{aze} & \textbf{sqi} & \textbf{som} & \textbf{sin} & \textbf{AVG} \\
\midrule
\multicolumn{10}{l}{\textbf{Accuracy}} \\
\textit{\quad Mistranslation} & -4 & -4 & -33 & -13 & -5 & -15 & -25 & -22 & \gooddown{-15.1} \\
\textit{\quad Overtranslation} & -6 & 19 & -7 & 13 & -2 & 2 & 18 & 7 & \badup{+5.5} \\
\textit{\quad Undertranslation} & 0 & -4 & -9 & 0 & 2 & -3 & -7 & -5 & \gooddown{-3.3} \\
\textit{\quad Addition} & 4 & 1 & 4 & 1 & -5 & -3 & 2 & -4 & \neutral{0.0} \\
\textit{\quad Omission} & 5 & 13 & 1 & 3 & 1 & 4 & -1 & 5 & \badup{+3.9} \\
\textit{\quad Untranslated} & -10 & 0 & -8 & -1 & -1 & -4 & -4 & -5 & \gooddown{-4.1} \\
\textbf{Total Accuracy} & \textbf{-11} & \textbf{25} & \textbf{-52} & \textbf{3} & \textbf{-10} & \textbf{-19} & \textbf{-17} & \textbf{-24} & \gooddown{-13.1} \\
\addlinespace
\multicolumn{10}{l}{\textbf{Fluency}} \\
\textit{\quad Grammar} & -1 & -4 & -8 & -7 & -1 & 5 & -14 & -6 & \gooddown{-4.5} \\
\textit{\quad Spelling} & -7 & -2 & 0 & -10 & -2 & 4 & 4 & -81 & \gooddown{-11.8} \\
\textit{\quad Punctuation} & -1 & -15 & -1 & -9 & -1 & 3 & 0 & -5 & \gooddown{-3.6} \\
\textbf{Total Fluency} & \textbf{-9} & \textbf{-21} & \textbf{-9} & \textbf{-26} & \textbf{-4} & \textbf{12} & \textbf{-10} & \textbf{-92} & \gooddown{-19.9} \\
\addlinespace
\multicolumn{10}{l}{\textbf{Style}} \\
\textit{\quad Awkward} & 3 & 12 & -11 & -2 & -7 & 11 & 10 & 0 & \badup{+2.0} \\
\textit{\quad Register} & 0 & 2 & -2 & -1 & -1 & -5 & 3 & -2 & \gooddown{-0.8} \\
\textbf{Total Style} & \textbf{3} & \textbf{14} & \textbf{-13} & \textbf{-3} & \textbf{-8} & \textbf{6} & \textbf{13} & \textbf{-2} & \badup{+1.3} \\
\addlinespace
\multicolumn{10}{l}{\textbf{Terminology}} \\
\textit{\quad Inconsistent} & -5 & 1 & 0 & 1 & 3 & -1 & 0 & 4 & \badup{+0.4} \\
\textit{\quad Wrong} & -8 & 1 & -10 & -7 & -7 & 0 & 6 & -12 & \gooddown{-4.6} \\
\textbf{Total Terminology} & \textbf{-13} & \textbf{2} & \textbf{-10} & \textbf{-6} & \textbf{-4} & \textbf{-1} & \textbf{6} & \textbf{-8} & \gooddown{-4.3} \\
\addlinespace
\textbf{Non-translation} & \textbf{-4} & \textbf{0} & \textbf{-10} & \textbf{-3} & \textbf{0} & \textbf{-1} & \textbf{-4} & \textbf{-1} & \gooddown{-2.9} \\
\bottomrule
\end{tabularx}
\captionof{table}{AutoMQM analysis comparing NLLB-54B and Gemini 2.5 Flash (sentence-level) on the OpenWHO test set.
\textbf{Top}: Overall MQM scores (higher is better). Gemini consistently outperforms NLLB.
\textbf{Bottom}: Difference in error counts (Gemini errors minus NLLB errors) per category. Negative values indicate Gemini made fewer errors for that category. Gemini outputs less major errors like mistranslations and incorrect terminology, at the cost of a slight increase in over-translation and omissions.}
\label{tab:automqm-openwho}
\end{minipage}

\clearpage

\clearpage

\section{Prompts}
\label{sec:appendix-prompts}

\begin{promptbox}
    \textbf{System:} Translate from English to [target lang name]. Give only the translation, and no extra commentary, or chattiness. Wrap the translated sentence in <result></result> tags.
    \vspace{1em}

    \textbf{User:} <text to translate>She lives in Boston.</text to translate>
    \vspace{1em}

    \textbf{Assistant:} <result>[Google Translate of ``She lives in Boston.'' into target lang]</result>
    \vspace{1em}

    \textbf{User:} <text to translate>[sentence to translate]</text to translate>
\end{promptbox}
\begin{promptcaption}
    Prompt used for \textbf{Sentence level} translation. We ask the model to wrap the translation in <result> tags to avoid model commentary interfering with translation accuracy measurement.
\end{promptcaption}

\begin{promptbox}
    \textbf{System:} Using the provided context, translate the ``Sentence to translate'' from English to [target lang name]. Give only the sentence translation, and no extra commentary, or chattiness. Wrap the translated sentence in <result></result> tags.
    \vspace{1em}

    \textbf{User:} <context>\newline
Her name is Mary. She lives in Boston. She is a doctor.\newline
</context>

Sentence to translate:\newline
She lives in Boston.
    \vspace{1em}

    \textbf{Assistant:} <result>[Google Translate of ``She lives in Boston.'' into target lang]</result>
    \vspace{1em}

    \textbf{User:} <context>\newline
[preceding sentence][sentence to translate][next sentence]\newline
</context>

Sentence to translate:\newline
[sentence to translate]
\end{promptbox}
\begin{promptcaption}
    Prompt used for \textbf{Sentence window} translation.
\end{promptcaption}

\begin{promptbox}
    \textbf{System:} Using the provided context, translate the ``Sentence to translate'' from English to [lang name]. Give only the sentence translation, and no extra commentary, or chattiness.
    \vspace{1em}

    \textbf{User:} <context>\newline
Her name is Mary. She lives in Boston. She is a doctor.\newline
</context>

Sentence to translate:\newline
She lives in Boston.
    \vspace{1em}

    \textbf{Assistant:} <result>[Google Translate of ``She lives in Boston.'' into target lang]</result>
    \vspace{1em}

    \textbf{User:} <context>\newline
[whole document for the sentence]\newline
</context>

Sentence to translate:\newline
[sentence to translate]
\end{promptbox}
\begin{promptcaption}
    Prompt used for \textbf{Sentence + doc context} translation.
\end{promptcaption}

\clearpage
\begin{promptbox}
    \textbf{System:} Translate from English to [lang name]. Give only the translation, and no extra commentary, or chattiness. Use the same formatting as the source text to translate, with one sentence per line. Enclose your translation in <result></result> tags.
    \vspace{1em}

    \textbf{User:} <text to translate>\newline
Her name is Mary.\newline
She lives in Boston.\newline
She is a doctor.\newline
</text to translate>
    \vspace{1em}

    \textbf{Assistant:} <result>\newline
[Google Translate of ``Her name is Mary.'' into target lang]\newline
[Google Translate of ``She lives in Boston.'' into target lang]\newline
[Google Translate of ``She is a doctor.'' into target lang]\newline
</result>
    \vspace{1em}

    \textbf{User:} <text to translate>\newline
[document sentence 1]\newline
[document sentence 2]\newline
...\newline
</text to translate>
\end{promptbox}
\begin{promptcaption}
    Prompt used for \textbf{Document level} translation.
\end{promptcaption}

\begin{promptbox}
    \textbf{System:} Translate from English to [lang name]. Give only the translation, and no extra commentary, or chattiness. Use the same formatting as the source text to translate, with one sentence per line. Enclose your translation in <result></result> tags.
    \vspace{1em}

    \textbf{User:} <text to translate>\newline
Her name is Mary.\newline
She lives in Boston.\newline
She is a doctor.\newline
</text to translate>
    \vspace{1em}

    \textbf{Assistant:} <result>\newline
[Google Translate of ``Her name is Mary.'' into target lang]\newline
[Google Translate of ``She lives in Boston.'' into target lang]\newline
[Google Translate of ``She is a doctor.'' into target lang]\newline
</result>
    \vspace{1em}

    \textbf{User:} <text to translate>\newline
[document sentence 1]\newline
[document sentence 2]\newline
...\newline
</text to translate>

    \textbf{Assistant:} [assistant response from above]

    \textbf{User:} Please translate again for a better version. Be particularly mindful of using the right script and tone, of adapting to context, and of translating each sentence faithfully.
    
<text to translate>[same as above]</text to translate>]
\end{promptbox}
\begin{promptcaption}
    Prompt used for \textbf{Doc-level + self-correct} translation.
\end{promptcaption}

% \begin{table}[ht]
%     \centering
%     \small
%     \begin{tabular}{l rr rr}
%         \toprule
%         & \multicolumn{2}{c}{\textbf{ChrF} $\uparrow$} & \multicolumn{2}{c}{\textbf{MetricX} $\downarrow$} \\
%         \cmidrule(lr){2-3} \cmidrule(lr){4-5}
%         & \textbf{Sent-level} & \textbf{Doc-level} & \textbf{Sent-level} & \textbf{Doc-level} \\
%         \midrule
%         \multicolumn{5}{l}{\textbf{OpenWHO}} \\
%         \quad Gemini & 51.72 & 55.17 \goodup{3.44} & 3.03 & 3.08 \badup{0.05} \\
%         \quad DeepSeek & 47.14 & 48.88 \goodup{1.74} & 3.54 & 3.52 \gooddown{-0.02} \\
%         \quad Gemma3 & 47.76 & 47.37 \baddown{-0.39} & 3.90 & 4.17 \badup{0.27} \\
%         \midrule
%         \multicolumn{5}{l}{\textbf{WMT literary}} \\
%         \quad Gemini & 44.29 & 50.66 \goodup{6.37} & 4.94 & 3.76 \gooddown{-1.18} \\
%         \quad DeepSeek & 43.54 & 46.88 \goodup{3.34} & 5.36 & 4.57 \gooddown{-0.79} \\
%         \quad Gemma3 & 42.39 & 44.45 \goodup{2.06} & 5.40 & 5.26 \gooddown{-0.14} \\
%         \midrule
%         \multicolumn{5}{l}{\textbf{WMT news}} \\
%         \quad Gemini & 53.58 & 54.83 \goodup{1.24} & 2.77 & 2.69 \gooddown{-0.08} \\
%         \quad DeepSeek & 51.40 & 50.58 \baddown{-0.82} & 3.52 & 3.42 \gooddown{-0.11} \\
%         \quad Gemma3 & 50.71 & 50.57 \baddown{-0.14} & 3.61 & 3.74 \badup{0.13} \\
%         \bottomrule
%     \end{tabular}
%     \caption{Average document-level vs sentence-level difference on ChrF and MetricX.}
%     \label{tab:llm-doc-vs-sent-avg}
% \end{table}

\end{document}